# MACHINE LEARNING-BASED SPIN STRUCTURE DETECTION

**Authors**

Labrie-Boulay, Isaac*[1]; Winkler, Thomas B.*[1]; Franzen, Daniel[2]; Romanova, Alena[1]; Fangohr, Hans[3,4]; Kläui, Mathias[1]

1: Institute for Physics, Johannes Gutenberg University, Mainz 55099, Germany.

2: Institute of Computer Science, Johannes Gutenberg University, Mainz 55099, Germany.

3: Scientific Support Unit for Computational Science, Max Planck Institute for the Structure and Dynamics of Matter, Hamburg 22761, Germany.

4: Computational Modelling Group, Faculty of Engineering and Physical Sciences, University of Southampton, Southampton, Hampshire, United Kingdom.

*: Both authors contributed equally to the work

Mail: twinkler@uni-mainz.de, klaeui@uni-mainz.de,

**ABSTRACT**

One of the most important magnetic spin structure is the topologically stabilised skyrmion quasi-particle. Its interesting physical properties make them candidates for memory and efficient neuromorphic computation schemes. For the device operation, detection of the position, shape, and size of skyrmions is required and magnetic imaging is typically employed. A frequently used technique is magneto-optical Kerr microscopy where depending on the sample's material composition, temperature, material growing procedures, etc., the measurements suffer from noise, low-contrast, intensity gradients, or other optical artifacts. Conventional image analysis packages require manual treatment, and a more automatic solution is required.
We report a convolutional neural network specifically designed for segmentation problems to detect the position and shape of skyrmions in our measurements. The network is tuned using selected techniques to optimize predictions and in particular the number of detected classes is found to govern the performance. The results of this study shows that a well-trained network is a viable method of automating data pre-processing in magnetic microscopy. The approach is easily extendable to other spin structures and other magnetic imaging methods.

**INTRODUCTION**

In the last decade, magnetic quasiparticles [1], [2] have raised the interest of researchers due to their interesting properties and their potential applicability in next-generation memory and neuromorphic devices [2],[3]. In particular, skyrmions in magnetic thin-films are under investigation due to their topological stabilization, their small size, and their ability to be easily manipulated by spin currents [4], [5]. However, depending on the magnetic properties of the thin-film, the skyrmion size might vary from a few nanometres[6] up to a few micrometres [7]. Furthermore, skyrmions can be circular, elliptical or have other shapes and a typical spin structure of a skyrmion with a Bloch domain walls is shown in the inset of Fig. 2 (a). Skyrmions have been suggested and experimentally demonstrated for a wide range of non-conventional computing schemes, such as stochastic computing [8], [9], token-based Brownian computing [10], [11], and reservoir computing [12]–[14]. Common to realize all these approaches is the reliable detection of the skyrmions. In particular, magneto-optical Kerr effect (MOKE) microscopy is used to study magnetic skyrmions having a diameter up to the micrometre scale [5], [9], [9]. The obtained greyscale contrast image in a Kerr microscope corresponds to the out-of-plane magnetization of the sample. For the analysis of skyrmions and their static and dynamic properties one needs to identify the position of all skyrmions as well as their shape [9], [15]. This has been very labour intensive so far as it was done typically manually, precluding large-scale systematic analysis. And for using image analysis software manual data pre-processing is often necessary to be able to evaluate the data correctly. Challenges are thermal and electrical noise or low contrast, which might arise from a non-perfect alignment of the polarisation filters, stray light, but also from necessary non-magnetic capping layers to avoid the oxidation of the sample, while at the same time absorbing the light partially. Furthermore, an intensity gradient might be visible in the data If the sample is not perfectly aligned or inhomogeneous. Structural defects and or dust particles might also impede data evaluation as well as sample edges. In Fig. 2a, 2d, and 2g we show a collection of typical MOKE microscopy data.

The study of skyrmions typically involves evaluating the size, position, and tracing the motion of skyrmions. This is often done by first creating a binary mask of measurements and then subjecting it to standard data analysis techniques. Binary mask creation through image processing software requires manually performing a series of manipulations that are highly dependent on the measurement to convert the raw data to masks [16]. Some of these manipulations include denoising, contrast enhancement, the application of a Gaussian filter, thresholding, and more depending on the measurement. Furthermore, no classical filter can reliably distinguish skyrmions or different magnetic structures, such as stripe domains, vortices, or domain walls. Fig. 2d and 2g are examples of measurements that require significant effort to pre-process using conventional methods.

Neural networks can provide a powerful means of automating the evaluation process since image segmentation is one of the key abilities of convolutional neural networks. Huge progress has been made in the last decade due to the extended use of general-purpose graphical processing units[17] (GPGPUs). Since machine learning has previously been successfully applied to magnetic problems [18]–[21], we here tackle the key problem of detecting the most important spin structure, skyrmions, to automate the analysis of magnetic microscopy measurements to enable applications such as unconventional computing.

In this study, we explore the ability of a convolutional neural network (CNN) to segment our MOKE microscopy data and obtain an automatic detection of skyrmions. We compare a 2-class and a 3-class network including defects and we tune the CNN to optimize the skyrmion and defect segmentation that exceeds the performance of conventional pattern detection. So far, no machine learning-based approach has been proposed to tackle the specific task of segmentation of magnetic thin film spin structure data, however CNNs are a possible choice that needs to be explored as done in the following.

**RESULTS AND DISCUSSION**

Convolutional neural networks used for optical pattern recognition use both a down-sampling structure called the contraction path and an up-sampling structure called the expansion path (See Fig. 1). The contracting path consists of two 3x3 convolutions followed by a rectified linear unit (ReLU) activation layer, and the down-sampling is finally achieved using a 2x2 max pooling layer with a stride of 2 [21] (See Fig. 1) in our case. For every down-sampling block, the number of feature channels is doubled. The expansive path concatenates the corresponding depth feature map from the contracting path, applies two 3x3 convolutions and a ReLU activation layer, and up-samples the feature map via a 2x2 up-convolution [21] (See Fig. 1). The concatenation of high-resolution features from the left side and up-sampled output from the right side is required for localization. These are called shortcuts. It should also be noted that the frequently used U-Net implementation employs unpadded convolutions which has the effect of creating masks having reduced size relative to the input image. Our implementation uses consistent padding to keep the dimensions of the original image at the output. An illustration of the CNN used in this study can be seen in Fig. 1. This 2-class CNN has a dropout rate of 5% between its 3x3 convolution blocks. Dropout is a regularization technique that makes it so that nodes in a layer are randomly ignored throughout training [22], [23]. Details of the network and the analysis are provided in the Methods section.

To train our CNN we carried out 5 trainings over 100 epochs. We used a few TensorFlow *callback* functions to prevent overfitting (EarlyStopping, ReduleLROnPlateau, ModelCheckpoint [24]). The average MCC of the validation data prediction was 0.7520 with a standard deviation of 0.032. Examples of predictions for this benchmark model can be seen below in Fig. 2c, 2f, and 2i.

We note that the model has difficulties when classifying defects, see Fig 2f for an example. Often these defects have a wide mixture of very different contrasts caused by sample drift and background subtraction when taking MOKE microscopy measurements. This causes part of it to be classified as a skyrmion and the other as background. Even though defects are quite different in shape and contrast to the skyrmions, the models cannot easily classify the two as two distinct objects.

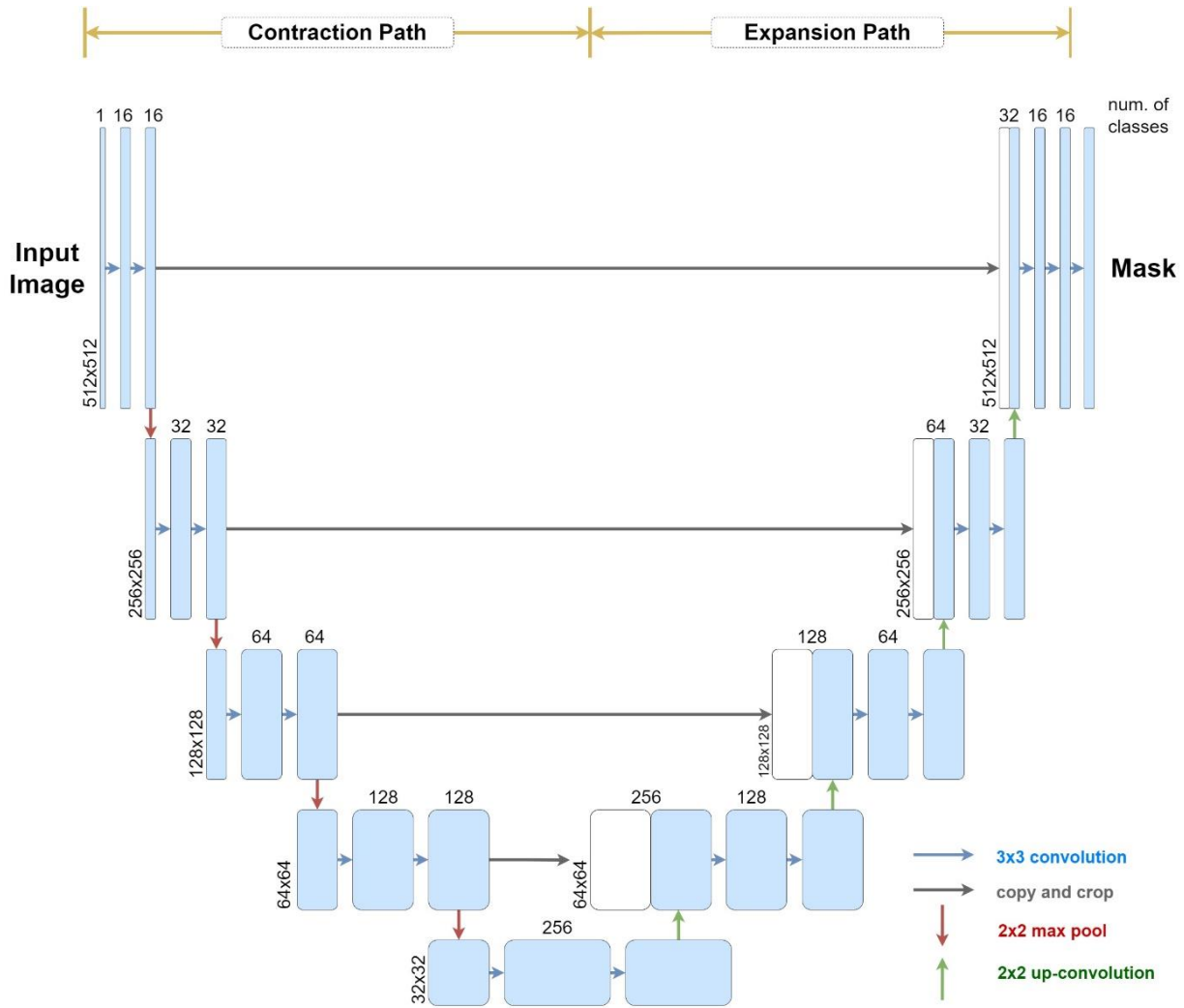

FIG.1. Network topology for the CNN used in this study. The left side is the contracting path and the right side is the expansive path. The grey arrows connecting both paths indicate the skip connections. Figure is adapted from Ref. [21].

Furthermore, in the case of noisy images, the masks created by the models are predicted to have tiny skyrmion speckles. Fig. 2i shows a prediction done on one such sample. This is problematic from an applied physics viewpoint. An experimentalist's analysis might, for example, involve finding and tracking the average size of skyrmions in a set of measurements. One can see how such small speckles might create difficulties for reliable detection.

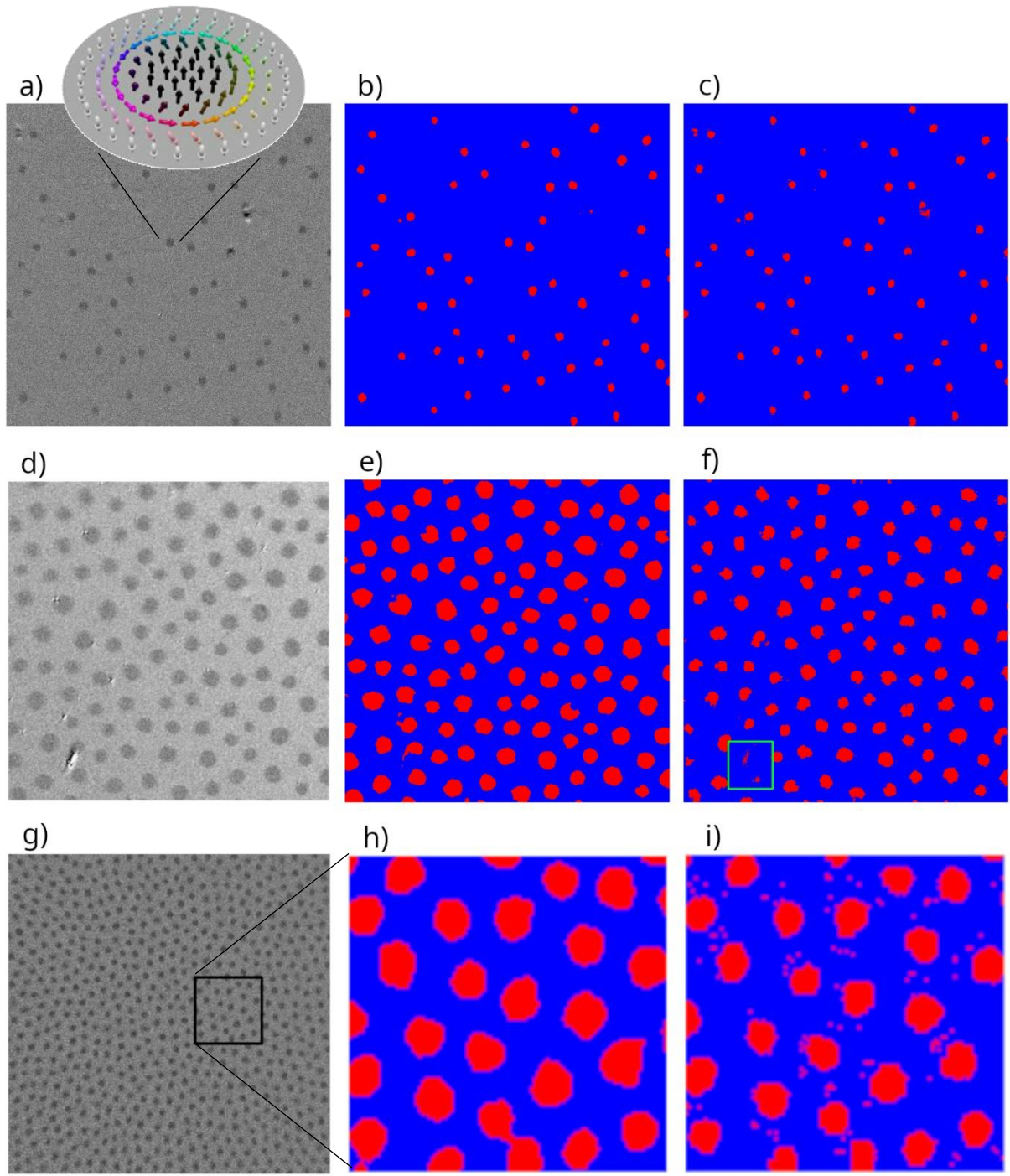

FIG 2. 2-Class Model Predictions. The pictures on the left show MOKE measurements which are difficult to mask using traditional methods, the middle column shows the labels of these measurements, and the right column shows predictions made by these 2-class models. (a) A sample having a few defects. The inset visualizes the spin structure of a skyrmion. The centre of the skyrmion with the magnetization "down" leads to black contrast and the surrounding area with the magnetization "up" leads to light grey contrast. Note that the coloured Bloch domain wall in the inset is not visible in the experimental image due to the limited spatial resolution. (b) Label of Fig. 2a. (c) Prediction of Fig. 2a. (d) This sample has defects and a slight brightness gradient. (e) Label of Fig. 2d. (f) Prediction of Fig. 2d where the green square indicates an example of how these models predict defects. (g) This sample is particularly noisy and the black square indicates a zoom area. (h) A 4x zoom on the label of Fig. 2g. (i) A 4x zoom on a prediction of Fig. 2g (note the small speckles indicating wrong detection).

In the spirit of interpretable AI, it was found that the cause for this problem is that the models are mostly predicting based on intensity values. We find that greyscale image analysis is able to reveal this: We tasked one of these models with making predictions on 255 uniformly coloured 8-bit greyscale images whose colour ranged from 0 to 255. This experiment found that any pixel having a value smaller than 88 was classified as skyrmion and any value above was predicted as background. The result can be seen in Fig. 3c. This provides clear evidence for this conjecture as to why the speckles appear in noisy images. Their noise produces a wide range of pixel brightness values. Random higher brightness pixels are classified as skyrmions.

Taking this analysis further, the model was tasked with predicting the inverted sample in Fig. 3a, where skyrmion pixels now have high greyscale values and background has low greyscale values. The model's prediction can be seen below in Fig. 3b. As expected, one can see that the model automatically classifies any dark object as skyrmion (white) which results in a nearly inversed prediction.

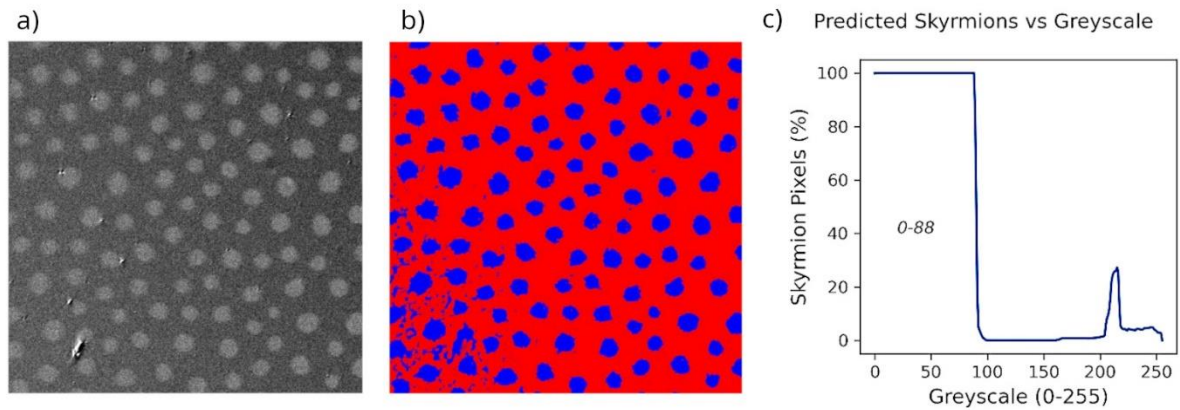

FIG. 3. Inverted Data Predictions. (a) Colour-inverted sample. (b) The prediction. This figure shows that any dark pixel is nearly automatically classified as skyrmion for the two-class case. The splash in the bottom left corner is likely caused by the fact that the sample has a colour gradient where the image gets progressively darker from the bottom left to the top right corner. (c) Greyscale pixels ranging from 0 to 88 are predicted as skyrmions 100% of the time.

To remove this shortcoming and create a model that is better suited for discriminating against defects and random noise, a third-class representing defects was added. By forcing the model to differentiate between the two different objects, the network is forced to learn to predict based on shapes and other factors beyond simply colour intensity. This new multiclass model (skyrmion, defect, background) was supervised using 3-class labels.

The training data for this work comprises data from previous publications, where areas with few defects were studied. Therefore defects in this data are rare. So here we add 100 new measurements deliberately containing many defects that were properly labelled by hand and a model was then trained on these images to be able to label defects on the original 1901 images. The predicted labels were post-processed so that the three-class model has 2001 images and is said to be weakly labelled. The three-class model also has a validation set and a testing set, which both contain 200 images/labels, and both labelled manually.

The MCC was computed by keeping skyrmions as the positive class and combining defects and backgrounds as the negative class. Just like the original network, the three-class CNN was trained 5 times over 100 epochs. The average MCC of the three-class predictions on the validation set was 0.849 with a standard deviation of 0.0075 which reveals a significant increase in predictive capability. Example predictions from this model can be seen in Fig. 4c, 4f, and 4i. It should be noted that now the small skyrmion speckles are all gone in the model.

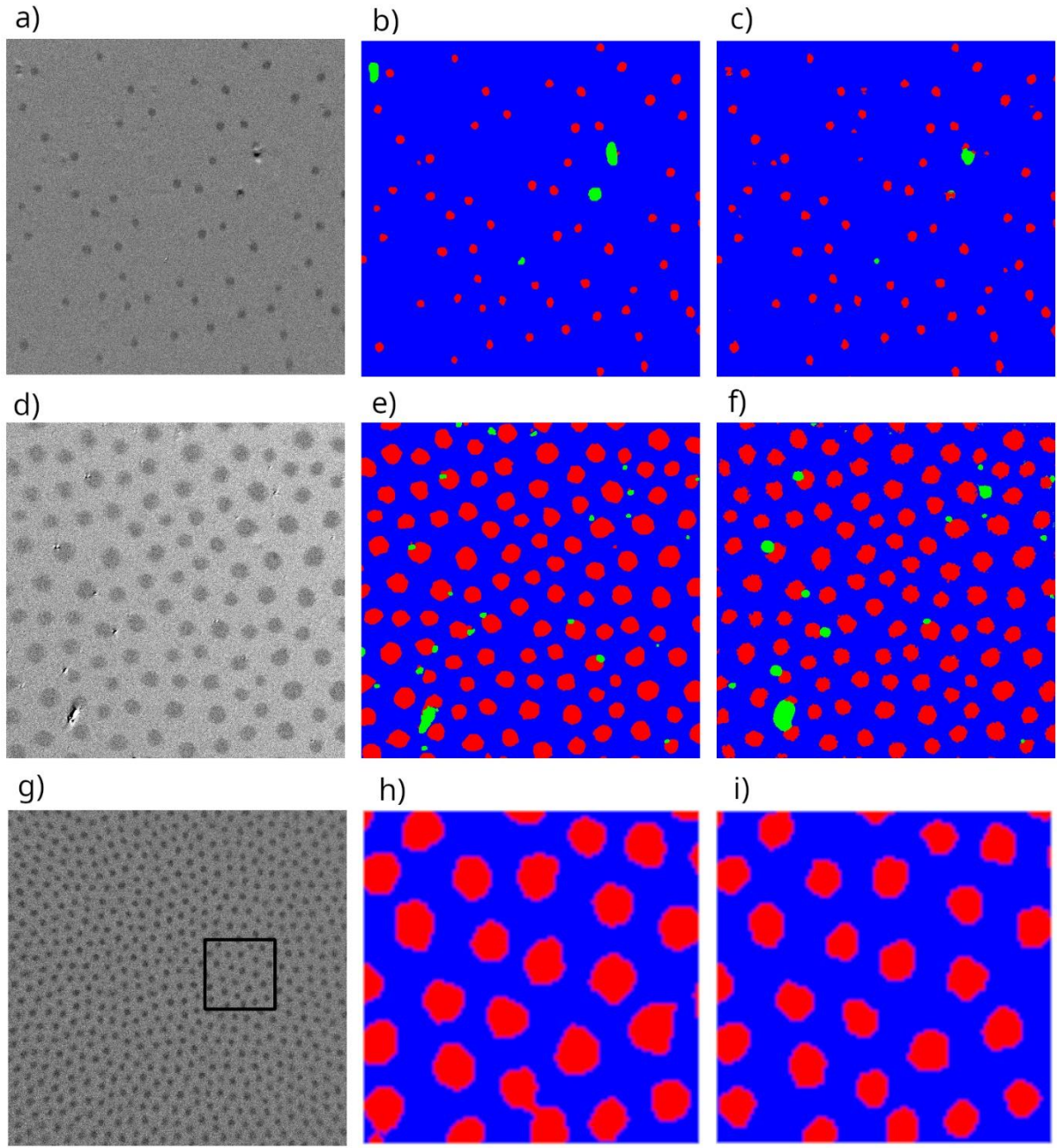

FIG 4. 3-Class Model Predictions. The pictures on the left show MOKE measurements, the middle column shows the labels (note the defects labelled in green), and the right column shows predictions made by these 3-class models. Figures (c) and (d) show that this model architecture can tell the difference between skyrmions and defects. As can be noted in (i), this kind of model no longer predicts small skyrmion speckles in noisy images.

Since the three-class model performs clearly better, the next step was to test additional further optimizations to the three-class neural network and to see if they improve performance further. One modification is made at a time and the mean and standard deviation of the MCC is recorded. In the end, all the modifications showing improved performance were combined to generate a master model. The master model was then tested against the test data to prevent bias towards the validation set. For this study, the training epochs were reduced and restricted to 15. The callback function hyperparameters were also adjusted accordingly.

The optimizations tested on our network and their resulting performance can be seen in Table 2. It should be noted that both the 2-class and 3-class benchmark models were also trained over 15 epochs to compare them with the optimized models. We explored different activation functions, fine tuning the dropout percentage, different levels of label smoothing, and two different data augmentation schemes.

Data augmentation refers to a technique that performs random transformation on the data during training. In that way, the number of training data is artificially increased, as previously proposed [25]. Test time augmentation refers to a data augmentation that happens at the prediction stage (after training). It entails pooling predictions from several transformed versions of a given test input to obtain a "smoothed" prediction [26]. In our case, a prediction is made on three differently transformed versions of the image, and the final prediction is averaged over the three.

The data transformations used in the data augmentation can be seen below in Table 1. Applying random 90° rotations, random lateral shifts, or a random scaling simply artificially expands the amount of measurement images seen. The maximum scaling factor of 0.2 is chosen, so the model is not trained on images containing skyrmions that are unnecessarily big. Some images are randomly injected with Gaussian noise to give the model access to more noisy data. A random change in brightness and contrast is also applied to images. Because of the nature of the MOKE microscopy measurements, some of the defects contain pixels that are clipped or close to being clipped by the 255 limit of 8-bit greyscale images. The high brightness of defects is caused by the continuous background subtraction from the Kerr microscope measurements. These background subtractions are essential to obtain magnetic data but after a certain time the shifting of the microscope illumination intensity becomes so significant that it creates shadow-like shapes around very bright defects. To prevent the model to register objects that are uniformly composed of clipped pixels, the brightness and contrast are only allowed to be amplified by a factor of $\leq 0.2$ as seen again in Table 1.

A subset of models were trained using both the same data augmentation techniques and random image inversion. The idea was to force the network to distinguish skyrmions not only by relative contrast differences but by shape in general. It should be noted that an inverted magnetic field creates negative images in MOKE measurements, therefore, making this potential feature very valuable.

| Data Augmentation Techniques |
| --- |
| Random 90° rotation |
| Gaussian noise |
| Random shift |
| Random scaling (up to a maximum factor of 0.2) |
| Random contrast change with a (-0.3, 0.3) limit |
| Random brightness change with a (-0.3, 3) limit |
| Image Inversion* |

TABLE 1. Data Augmentation Used. This table summarizes the data transformations used in the data-augmented models. Image Inversion is marked with an asterisk to signify that only a subset of models was subjected to this augmentation both in training and at test time.

| Optimization Type | Optimization | Mean & Standard Deviation (MCC) |
|---|---|---|
| 2-class model | N/A | 0.752 (SD = 0.064) |
| 3-class initial model | N/A | 0.896 (SD = 0.002) |
| Data Augmentation | w/o inversion | **0.898 (SD = 0.003)** |
| | w inversion | 0.884 (SD = 0.012) |
| Activation Function | prelu | 0.885 (SD = 0.006) |
| | tanh | 0.869 (SD = 0.014) |
| | mish | 0.891 (SD = 0.006) |
| Label Smoothing | 30% | 0.875 (SD = 0.033) |
| | 40% | 0.884 (SD = 0.005) |
| Dropout | 0% | 0.896 (SD = 0.002) |
| | 10% | **0.900 (SD = 0.006)** |
| | 15% | 0.892 (SD = 0.004) |

TABLE 2. Performance of three-class model optimizations. This table shows the MCC value for each optimization. The tested activation functions were parametric rectified linear unit function (PReLU), a hyperbolic tangent function (Tanh), and "Mish", mathematically defined as $f(x) = x \tanh(softplus(x))$. All are common activation functions used in neural networks. For label smoothing, we explored with $\alpha = 0.3$ and $\alpha = 0.4$. The default model uses $\alpha = 0.2$. We also tried training the model using different amount of dropout following the 3x3 convolutions (10% and 15%).

Considering the standard deviations shown in Table 2, none of the optimizations show very large improvements relative to the 3-class initial benchmark. Most resulted in worse performance. Our default model's activation function (ReLu) and label smoothing (20%) performed the best. Data augmentation improved performance by an MCC increase of 0.002. Data augmentation with image inversion most likely results in the models having to solve the much more sophisticated task of perfectly recognizing the shape of skyrmions and defects hence the reduction of performance. Also, increasing the dropout rate to 10% seems to increase the MCC by 0.004. Taking into account these slight improvements, the master network was created.

Finally, the master model was trained for 15 epochs and tasked with predicting the test set. The prediction results for all data sets can be seen in Fig. 5. The validation and training set performances can also be seen in this figure. The master models show a nearly identical performance to the benchmark models when predicting the test set. The generated masks are all of reasonable quality.

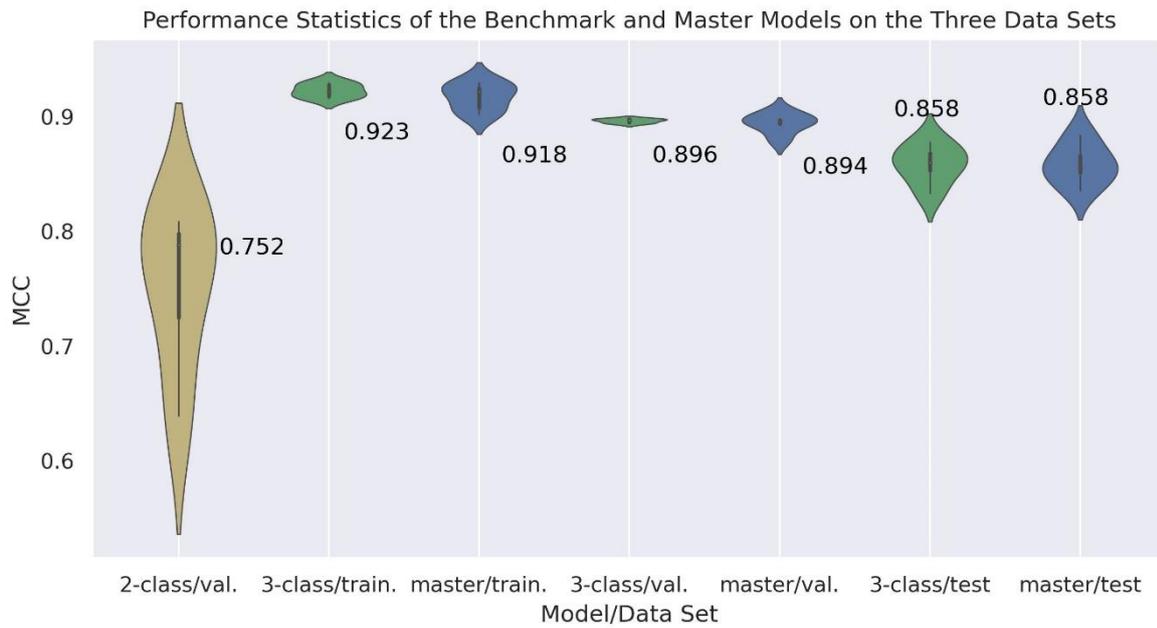

FIG 5. Performance Statistics of the Benchmark and Master Models on the Three Sets. Plot comparing the average MCC and MCC standard deviation of the 3-class benchmark model and the master model on all three data sets.

The MCC metric is only a pixel-wise quality measure for the prediction. Since we are interested in whole objects we compared the true size of the skyrmions in the test set to the predicted size of the skyrmions. This way, the investigation is semantic rather than at the pixel-level. A histogram comparing the true size of the skyrmions to the master model's predictions of the sizes is shown in Fig. 6. As one can see, the prediction's distribution matches the true size distribution. However, a second smaller peak at around 150 pixels can be seen. This being said, because of the difficulty of estimating the real size of skyrmions when labelling the data and as a result of the weak relabelling of the 3-class labels, the size of skyrmions in the labels should not be confused with the true size of the skyrmions. In other words, the labels themselves do not perfectly represent the magnetization profile.

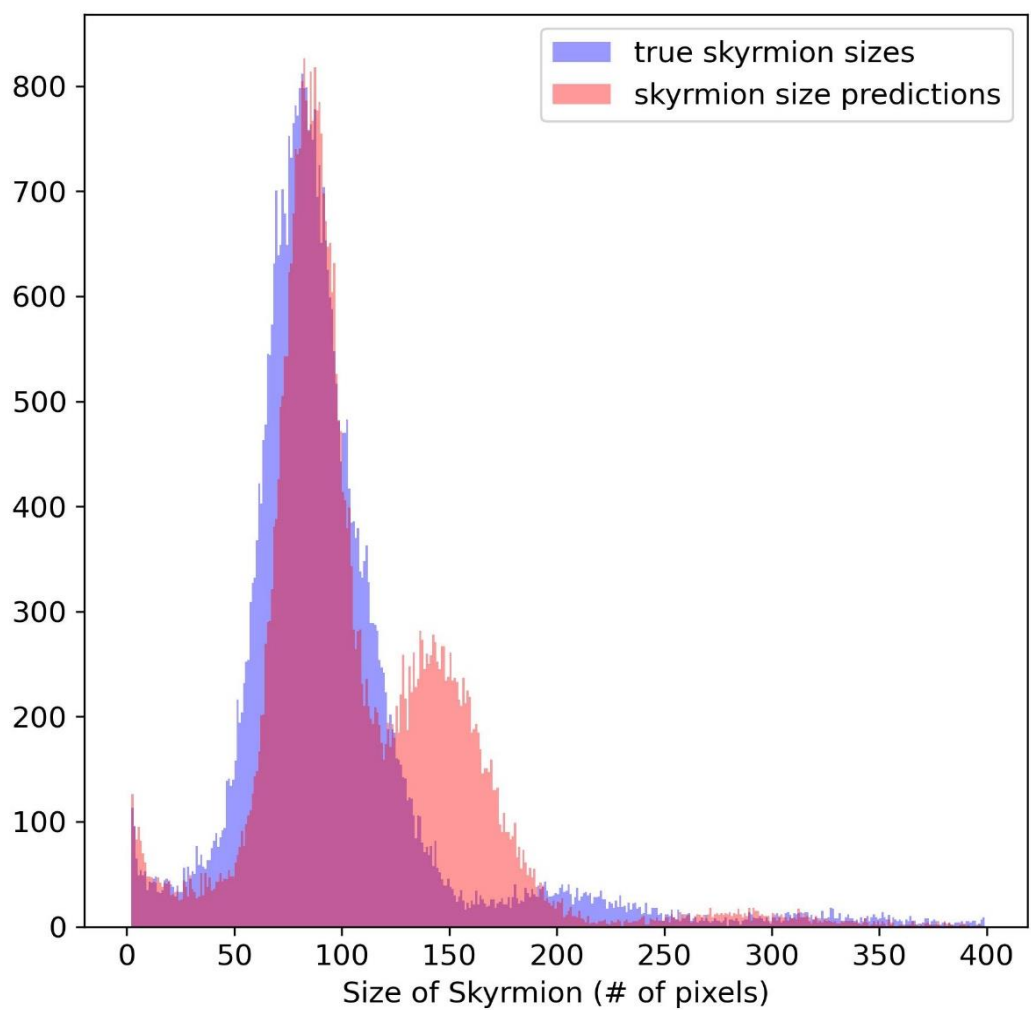

FIG. 6. Predictions on the Size of the Skyrmions in the Test Set. This figure shows the skyrmion size distributions of the test set skyrmion labels and the master model predictions on this test set.

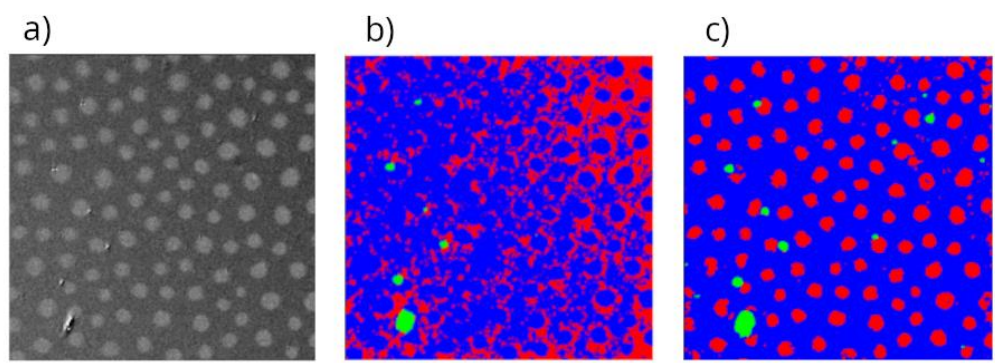

FIG 7. Inverted data predictions from the master model and model trained on inversion data augmentations. (a) The inverted sample. (b) The master model prediction. (c) The inversion model predictions.

**CONCLUSION**

In summary, it was shown that our optimized convolutional neural network can produce machine-learning models which create accurate skyrmion detection masks from magnetic Kerr microscopy images. Binary classifiers have difficulties differentiating between defects and skyrmions and do not perform very well on noisy images. We have found that a reason for this is that the 2-class model relies too much on brightness/contrast when making predictions. Adding defects as a third class drastically improves the quality of the prediction. The average MCC significantly improves and the predictions are very close to ground truth. Datasets and networks are uploaded for open-access use [27].

The finetuning of the network to create a master convolutional neural network showed no strongly improved performance when tested against the test set. This being stated, none of the individual optimizations showed clear individual impact for an improvement. The reason for this is that most of our 3-class models can easily perform this task and that it may be extremely difficult to break an MCC greater than 0.90 given that most predictions almost flawlessly passed a visual inspection test already before the finetuning. This could be due to the training data set being very strong or that the test set is lacking in more difficult samples. For this study, the dataset remained limited to data obtained from previous experiments. For now, to challenge the network's predictive power beyond the test set, we tested a model trained with image inversion data augmentation on an inverted image, which also performed reasonably well, see Fig. 7. Still, the master model in general outperformed the model's trained with the inversion augmentation. Maybe a less accurate but more modular model able to do inverted images is in fact more desirable for daily use. Also, when magnetic contrast switches within a measurement, the ability to detect both configurations with only one inference may be convenient.

To conclude, we have developed a convolutional neural network to detect and characterize skyrmions in magnetic microscopy images. As pre-processing noisy Kerr microscopy skyrmion images for experimental physics is complex and time-consuming, therefore automatization is important to increase the efficiency of the research. This paper demonstrates that a CNN can be trained to create very accurate skyrmion masks given the appropriate training data. A possible future avenue for this research would be to see if a neural network can be trained to automate the pre-processing of other magnetic structures such as skyrmioniums [28], stripe domains [28], or even domain walls [28]. Our approach is furthermore applicable to many magnetic microscopy techniques beyond Kerr microscopy, such as magnetic force microscopy, Lorentz microscopy or Scanning electron microscopy with polarization analysis making our results broadly applicable. Furthermore, one could see if more powerful models can be created if trained with videos instead of images (with Long-Short Term Memory or Echo-State networks) [29], [30]. We provide data and trained models on zenodo, see Ref. [27] .

**METHODS**

The labels of the data sets are one-hot encoded for two classes. The model was trained on 1901 training images sampled from 19 separate Kerr microscope videos. The labels were manually created using conventional mask creation techniques with the image processing software ImageJ[16]. The validation data set is composed of 200 images taken from 5 different videos. Different convolutional neural networks can be used. Here we have chosen a U-Net derivative as the framework is flexible for the detection of different physical entities in images [21]. After training, the model generates a prediction on the validation set and from there one can extract its performance metric based on a confusion matrix. The performance metric serves as a method for comparing models when one attempts to optimize a neural network for a specific task. After all optimizations have been applied, the test set is used for a final benchmarking. In our case, the test set is composed of 200 images coming from 6 different videos taken on different samples. It is important to note that both the validation set and the test set include images of skyrmions of all sizes. It includes images with different degrees of noise, and images with different defects. The training set has a total of 1466843 skyrmions and 18.9% of its pixels are labelled as skyrmions. The validation set has 17.8% skyrmions and the test set has 9.9%.

The score used for benchmarking our models is the Matthews Correlation Coefficient (MCC) [31]. Studies have shown that the MCC gives a strongly indicative performance score on both positively-biased and negatively-biased predictions for every kind of dataset [31]. Generally speaking, a positively-biased prediction accurately predicts positive values but does poorly when predicting negative ones. The MCC has the ability to properly score predictions on positively and negatively-biased predictions on positively imbalanced datasets, negatively imbalanced datasets, and balanced datasets [31]. As such, this metric is used as a reference performance measure for unbalanced data in different fields of research. In our dataset, skyrmions are classified as positive and background as negative. Our datasets are negatively imbalanced as there are more background pixels than skyrmion pixels in the images. Therefore, the MCC's ability to account for class imbalance is very important in our case.

The MCC for a binary classifier is given by,

$$MCC = \frac{TP \times TN - FP \times FN}{\sqrt{(TP+FP)(TP+FN)(TN+FP)(TN+FN)}}$$ [31].

The training is run five times for every model. The MCC is calculated for every run. The mean and standard deviation of each model's performance are then calculated. This method of analysing the performance of a model gives a more realistic idea of the performance. Note that, unlike other standard measures, the $MCC \in [-1,1]$ where -1 refers to a perfectly incorrect classification, 0 means perfectly inconclusive, and 1 indicates a perfect classification. The MCC however only measures the prediction performance in a pixel-wise approach but lacks physical interpretation of the data.

The models were written in python using the TensorFlow 2.9 [24] machine learning library with Keras [32] running on top providing the high-level API. This study's software also utilized functionalities from numpy [33], scikit-learn [34], and PIL [35]. All models were trained on an Nvidia RTX 3060 GPU [17].

**ACKNOWLEDGMENTS**

I. Labrie-Boulay, gratefully acknowledges the German Academic Exchange Service (DAAD) and Mitacs for awarding him the scholarship that allowed him to complete his internship held at the Institut für Physik at the Johannes Gutenberg University of Mainz. He also thanks T. B. Winkler for providing him with technical assistance and supervising him throughout his internship. We thank Karin Everschor-Sitte for the fruitful discussions. Daniel Franzen was funden by DFG Project number: 233630050 (TRR 146). The work was further funded by the emergentAI center, funded itself by the CarlZeissStiftung, further by the Deutsche Forschungsgemeinschaft (DFG, German Research Foundation) projects 403502522 (SPP 2137 Skyrmionics), 49741853, and 268565370 (SFB TRR173 projects A01 and B02). The work is also supported by the Horizon 2020 Framework Program of the European Commission under FET-Open grant agreement ERC-2019-SyG no. 856538 (3D MAGiC) and the Horizon Europe project no. 101070290 (NIMFEIA), which we acknowledge.